\documentclass{article} 
\usepackage{iclr2018_conference,times}
\usepackage{hyperref}
\usepackage{url}
\usepackage[utf8]{inputenc} 
\usepackage[T1]{fontenc}    
\usepackage{hyperref}       
\usepackage{url}            
\usepackage{booktabs}       
\usepackage{amsfonts}       
\usepackage{nicefrac}       
\usepackage{microtype}      
\usepackage{graphicx}
\usepackage{subfigure}
\usepackage{amsmath}
\usepackage{algorithmic}
\usepackage{algorithm}
\usepackage{amssymb}
\iclrfinalcopy

\title{Deep Active Learning over the Long Tail}

\author{Yonatan Geifman\\
	Computer Science Department\\
	Technion – Israel Institute of Technology\\
	\texttt{yonatan.g@cs.technion.ac.il} \\
	\And
	Ran El-Yaniv\\
	Computer Science Department\\
	Technion – Israel Institute of Technology\\
	\texttt{rani@cs.technion.ac.il} \\
}

\begin{document}

\maketitle

\begin{abstract}
This paper is concerned with pool-based active learning for deep neural networks. 
Motivated by coreset dataset compression ideas, we present a novel active learning algorithm that queries consecutive points from the pool using farthest-first traversals in the space of neural activation over a representation layer.
We show consistent and overwhelming improvement in sample complexity over passive learning (random sampling) for three datasets: MNIST, CIFAR-10, and CIFAR-100. In addition,
our algorithm outperforms the traditional uncertainty sampling technique (obtained using softmax activations), and we identify cases where uncertainty sampling is only slightly better than random sampling.
\end{abstract}

\newtheorem{theorem}{Theorem}[section]
\newtheorem{lemma}[theorem]{Lemma}
\newtheorem{corollary}[theorem]{Corollary}
\newtheorem{observation}[theorem]{Observation}

\newtheorem{assumption}[theorem]{Assumption}
\newtheorem{example}[theorem]{Example}

\newenvironment{proofsketch}[1][Proof Sketch:]{\begin{trivlist} \item[\hskip \labelsep
		{\bfseries #1}]}{\hfill{$\Box$}\end{trivlist}}

\newcommand{\argmax}{\operatornamewithlimits{argmax}}
\newcommand{\argmin}{\operatornamewithlimits{argmin}}

\newcommand{\rnote}[1]{ {\color{red}  [ Ran: \textit{#1}]} }
\newcommand{\ynote}[1]{ {\color{blue} [ Yonatan: \textit{#1}]} }
\newcommand{\sharon}[1]{ {\color{blue} {#1}} }

\newcommand{\vect}[1]{\mathbf{#1}}
\newcommand{\one}{\mathbf{1}}

\newcommand{\hr}{\hat{r}}
\newcommand{\hphi}{\hat{\phi}}

\newcommand{\VS}{\mathcal{VS}} 

\newcommand{\data}[1]{\texttt{#1}}
\newcommand{\tdata}[1]{\scriptsize{\texttt{#1}}}
\newcommand{\COIL}{\data{COIL}}
\newcommand{\MUSH}{\data{MUSH}}
\newcommand{\MUSK}{\data{MUSK}}
\newcommand{\PIMA}{\data{PIMA}}
\newcommand{\BUPA}{\data{BUPA}}
\newcommand{\VOTING}{\data{VOTING}}
\newcommand{\MONK}{\data{MONK}}
\newcommand{\IONO}{\data{IONOSPHERE}}
\newcommand{\TAE}{\data{TAE}}
\newcommand{\DIGIT}{\data{DIGIT}}
\newcommand{\TEXT}{\data{TEXT}}

\newcommand{\sCOIL}{\tdata{COIL}}
\newcommand{\sMUSH}{\tdata{MUSH}}
\newcommand{\sMUSK}{\tdata{MUSK}}
\newcommand{\sPIMA}{\tdata{PIMA}}
\newcommand{\sBUPA}{\tdata{BUPA}}
\newcommand{\sVOTING}{\tdata{VOTING}}
\newcommand{\sMONK}{\tdata{MONK}}
\newcommand{\sIONO}{\tdata{IONOSPHERE}}
\newcommand{\sTAE}{\tdata{TAE}}
\newcommand{\sDIGIT}{\tdata{DIGIT}}
\newcommand{\sTEXT}{\tdata{TEXT}}

\newcommand{\myalg}[1]{\texttt{#1}}
\newcommand{\mytalg}[1]{\scriptsize \texttt{#1}}
\newcommand{\SGT}{\myalg{SGT}}
\newcommand{\ZHU}{\myalg{GRMF}}
\newcommand{\BELKIN}{\myalg{SOFT}}
\newcommand{\SCHOLKOPF}{\myalg{CM}}
\newcommand{\EXPP}{\myalg{+EXPLORE}}
\newcommand{\sEXPP}{\mytalg{+EXPLORE}}
\newcommand{\sZHU}{\mytalg{STRICT}}
\newcommand{\sBELKIN}{\mytalg{SOFT}}
\newcommand{\sSCHOLKOPF}{\mytalg{RANGE}}
\newcommand{\sSGT}{\mytalg{SGT}}

\newcommand{\ENG}{\myalg{ENERGY}} 

\newcommand{\KNN}{\myalg{kNN}}

\newcommand{\comp}[1]{\small \texttt{#1}}
\newcommand{\QEXPLORE}{\comp{Q-EXPLORE}}
\newcommand{\QEXPLOIT}{\comp{Q-EXPLOIT}}
\newcommand{\EXPPSWITCH}{\comp{EXPLORE-EXPLOIT-SWITCH}}

\newcommand{\cA}{{\cal A}}
\newcommand{\cB}{{\cal B}}
\newcommand{\cF}{{\cal F}}
\newcommand{\cG}{{\cal G}}
\newcommand{\cH}{{\cal H}}
\newcommand{\cL}{{\cal L}}
\newcommand{\cV}{{\cal V}}
\newcommand{\cX}{{\cal X}}
\newcommand{\hL}{{\widehat{L}}}
\newcommand{\hcL}{{\widehat{\mathcal{L}}}}
\newcommand{\hR}{\widehat{R}}
\newcommand{\bB}{\mathbf{B}}
\newcommand{\bW}{\mathbf{W}}
\newcommand{\bR}{\mathbf{R}}
\newcommand{\bD}{\mathbf{D}}
\newcommand{\bG}{\mathbf{G}}
\newcommand{\bL}{\mathbf{L}}
\newcommand{\bI}{\mathbf{I}}
\newcommand{\bC}{\mathbf{C}}
\newcommand{\bZ}{\mathbf{Z}}
\newcommand{\ba}{\mathbf{a}}
\newcommand{\bd}{\mathbf{d}}
\newcommand{\be}{\mathbf{e}}
\newcommand{\bg}{\mathbf{g}}
\newcommand{\Bf}{\mathbf{f}}
\newcommand{\Bg}{\mathbf{g}}
\newcommand{\bh}{\mathbf{h}}
\newcommand{\bp}{\mathbf{p}}
\newcommand{\bq}{\mathbf{q}}
\newcommand{\bt}{\mathbf{t}}
\newcommand{\bu}{\mathbf{u}}
\newcommand{\bv}{\mathbf{v}}
\newcommand{\bx}{\mathbf{x}}
\newcommand{\by}{\mathbf{y}}
\newcommand{\bz}{\mathbf{z}}
\newcommand{\E}{\mathbf{E}}
\newcommand{\var}{\text{var}}
\renewcommand{\H}{\mathbf{H}}
\renewcommand{\S}{\mathbf{S}}
\newcommand{\T}{\mathbf{T}}
\newcommand{\CM}{\mathtt{CM}}
\newcommand{\CMRAD}{\mathtt{CM-SUP}}

\newcommand{\eqdef}{\mathrel{\ensurestackMath{\stackon[1pt]{=}{\scriptstyle\Delta}}}}
\newcommand{\nchoosek}[2]{\left(\begin{array}{c}#1\\#2\end{array}\right)}
\renewcommand{\Pr}{\mathbf{Pr}}
\newcommand{\head}{\mathrm{head}}
\newcommand{\tail}{\mathrm{tail}}
\newcommand{\rad}{\mathtt{R}}
\newcommand{\parr}{\mathtt{Par}}
\newcommand{\tA}{\mathtt{A}}
\newcommand{\bpi}{\pmb{\pi}}
\newcommand{\cN}{{\mathcal N}}
\newcommand{\cY}{{\mathcal Y}}
\newcommand{\rE}{\mathbf{E}}
\newcommand{\pig}{\tilde\pi_{\gamma/k}}
\newcommand{\al}{\alpha}
\newcommand{\QED}{\hfill{$\Box$}}
\newcommand{\lam}{\lambda}
\newcommand{\err}{\mathop{\rm er}}

\newcommand{\I}{\mathbb{I}}
\newcommand{\hf}{f_Q}
\newcommand{\hg}{\hat{g}}
\newcommand{\LA}{{\mathcal L }}
\newcommand{\GLi}{G_L^{(i)}}
\newcommand{\GUi}{G_U^{(i)}}
\newcommand{\fail}{\texttt{fail}}
\newcommand{\reals}{\mathbb{R}}

\section{Introduction and Related Work}
\label{sec:Intro}

\emph{Active learning} provides a learning algorithm with some control over the learning process, potentially leading to significantly more efficient learning in terms of labeling efforts \citep{CohAtlLad94}.
The  key question in active learning is how many label requests are sufficient to train a classifier to a specified accuracy, a quantity known as  \emph{label complexity} \citep{hannekestatistical}. 
Theoretically, there are instances where effective active learning can achieve
`exponential speedup' -- roughly meaning that to achieve $\epsilon$
excess-risk, the dominating factor in the active learner's label complexity will be $O(\log(1/\epsilon))$ rather than $O(1/\epsilon)$
(or $O(1/\epsilon^2)$ in the agnostic case) \citep{balcan2006agnostic}.
This huge asymptotic potential saving is tantalizing.
With the growth of deep and large neural models that are hungry for huge labeled samples, the importance and need for effective active learning techniques is only growing. 
Unfortunately, the literature on active learning in deep neural networks is extremely sparse, and the few existing works fall short of providing viable active learning solutions for practical applications. 

One major obstacle in applying any active learning algorithm
without prior knowledge, is the need to 
select its hyper-parameters on the fly, while acquiring labeled samples.
This is a challenging task, especially in the early stages of the active learning
session while the currently available labeled dataset is still small and extremely biased. 
Many seminal works on active learning avoided dealing with this hard problem altogether, by
intentionally selecting a good set of hyper-parameters based on `prior knowledge' \citep{tong2001support,baram2004online,gal2017deep}.
The few studies that included hyper-parameter selection
within the active learning process do not report significant improvements of the active learning algorithm over random sampling (``passive learning''); see, e.g. \cite{huang2015efficient}.
In the context of deep nets, because of the extreme sensitivity of large neural nets
to a variety of hyper-parameters (such as learning rates, initialization schemes, regularization techniques, and the architecture itself), this hyper-parameter selection trap is expected to be much more severe. We believe that 
this inherent difficulty hinders further developments of deep active networks, at least within the traditional setting of active learning.

In this paper we consider a ``long-tail'' variant of pool-based 
active learning. In our variant, a (deep) model has already been initially trained to achieve reasonable 
accuracy.
Then, in a sequence of phases, and based on budgetary constraints (or accuracy improvement requirements),
additional labels are sought to increase the model's 
accuracy. In each phase, a specified number $m$ of samples to be labeled are 
to be selected  from a given 
large pool of unlabeled samples. At this point, an active learning algorithm is applied 
and its performance should improve upon random sampling of the $m$ points from the pool, uniformly at random (i.e.,
passive learning). 
Thus, this active setting variant essentially differs from the standard pool-based 
setting \citep{tong2001support} in its starting point.
In our setting, we don't expect to gain anything during
the early stages of the active learning. On the contrary, we are willing to invest on random sampling at  
the beginning in order to gain much later when improving
the model, over the ``long tail'' of the training process.
Active learning, thus, becomes the means to expedite 
model improvements.
We show that our setting mitigates the hyper-parameter selection challenge, stabilizes the active performance (which is typically extremely noisy 
and unreliable in early active stages), and overall, allows for 
practical deep active learning with quite impressive sample complexity speedups,
compared to passive learning.

Our long-tail active learning setting and our new active learning algorithm (see Section~\ref{sec:active_deep}) are inspired by a novel
study on dataset compression described below, whereby the goal is to take a given deep model 
trained on some dataset and find, in hindsight, a subset of training samples that will generate a similar performance (when training the model over the compressed dataset).
We show that activation levels in the representation layer provide sufficient
information to compress CIFAR-10 to 50\% of its original size while compromising
only 1.5\% in accuracy. This level of dataset compression has not been previously 
reported. 
The ability
to compress a dataset using representation-layer activations of an
already trained model motivates both our active learning setting and our new active learning algorithm itself.

There is a rich literature on active learning, which is beyond the
scope of this paper. For a sample of the classic and some modern works, see 
\citep{CohAtlLad94,FreundSST93,tong2001support,baram2004online,balcan2013active,huang2015efficient}.
For a comprehensive summary of the theory of (disagreement-based) active learning, and its relationship to selective prediction,
see \citep{hannekestatistical,el2012active}. 
A recurring idea in (pool-based) active learning is 
that of \emph{uncertainty sampling}, which means that 
unlabeled pool of points are prioritized by model uncertainty,
and the most uncertain points should be queried first.
Applications of uncertainty sampling  depend on the
model. In neural networks uncertainty of a point $x$ is approximated 
by the network's \emph{Softmax Response} (SR) activation recorded for $x$, which reflects distance from the decision boundary.
In the context of deep nets, \citet{gal2017deep}
presented active learning algorithms based on a clever 
Monte-Carlo dropout technique and applied them on the MNIST dataset using a relatively small 
network, and for detecting skin cancer from images by
fine-tuning a pre-trained VGG16 architecture. 
\citet{wang2016cost} applied the well-known softmax response (SR)
idea supplemented with pseudo-labeling (self-labeling of highly confident 
points) for actively learning the `cross-age-celebrity' dataset and 
Caltech-256, using a
deep architecture that was pre-trained over Imagenet.
\cite{zhou2013active} constructed a deep architecture based on
a restricted Boltzmann machine (RBM) to actively learn sentiment categorization.
Their active learning algorithm relied on pre-training the RBM over a large unlabeled dataset,
and their querying function used SR. 
All these works achieved reasonable active learning performance by exploiting 
prior knowledge, which significantly helped to surmount the hyper-parameter and model selection obstacles. In this sense, these works strongly support and motivate the setting we propose here with ``late'' starting of the active queries.
In addition,  these works heavily relied on 
the softmax response idea, which 
was slightly improved upon by \citet{gal2017deep} using their Monte-Carlo dropout technique, 
which can be viewed as consolidating several 
independent applications of softmax response.



\subsection{Problem Setting}
\label{sec:setting}
Consider a standard supervised learning problem defined in terms of a feature space $\cX$, a label space $\cY$, and an  underlying distribution $P(X,Y)$, where $X \in \cX$, $Y \in \cY$. Based on a labeled training set $S_m =\{x_i,y_i\}$
of $m$ training samples, the goal is to select  
a prediction function $f\in \cal{F}$, $f:\cal{X}\rightarrow \cal{Y}$,
so as to minimize the risk $R_\ell(f)=\E_{(X,Y)}[\ell(f(x),y)]$,
where $\ell(\cdot) \in \reals^{+}$ is a given loss function. 

In our ``long-tail'' active learning setting, we assume 
that a ``reasonable'' function $f \in \cF$ has already been trained using supervised learning over the random sample $S_m$. We are also given a pool $U$ of $u \gg m$ unlabeled samples from $P(X)$. We now consider two problem variants:
\begin{enumerate}
\item \emph{Budget-constrained}: Given a budget $n$ for labeling,
actively select from $U$, $n$ unlabeled samples and request their label to obtain a labeled set $S_n$. Then use $S_m \cup S_n$ to train $f' \in \cF$
whose risk is the smallest possible.
\item \emph{Error-reduction}: Given an $\epsilon \in [0,1]$, select 
from $U$ the minimal number $n$ of samples, whose labels will be 
requested so as create the labeled sample $S_n$. Use 
$S_m \cup S_n$ to find $f' \in \cF$ such that
$R_{\ell}(f') \leq R_{\ell}(f) - \epsilon$.
We note that $\epsilon$ should be sufficiently small to enable this task to be accomplished (and we cannot know in advance if the required error-reduction is achievable).
\end{enumerate}

The performance of the ``passive'' solution for both these variants will be the natural benchmark for any active method. The passive
solution
is obtained by training $f'$ using a training set created by 
random (uniform) sampling of points in $U$. 
For example, in the budget-constrained variant, the passive learning algorithm samples a subset
of $n$ points from the given $u$ points uniformly at random to create $S_n$.

Finally, we decompose a deep neural model $f$ as $f(x)=\tau(\phi(x)):\cal{X}\rightarrow\cal{Y}$, where $\phi(\cdot)$ consists of the first part of the network from the input layer 
until (and including) a higher representation layer,
and $\tau(\cdot)$ represent the final layers. 
In this work we consider the representation layer 
as the second last layer and $\tau$ is the classifier in the last layer.


\section{Motivation: Compression Schemes for Deep Learning}
\label{sec:compression}

A \emph{coreset} is a sub-sample of a dataset, which can be used as a proxy for the full 
set. The idea in the study of coresets \citep{phillips2016coresets} is to 
use them to approximate the full dataset such that the output of 
an algorithm over the coreset will be qualitatively similar to its output over the full dataset (with respect to some cost function).
Coresets can be used to create efficient approximation algorithms
by running the same algorithm on a small fraction of the input data.
Many coresets ideas are based on computational geometry.
The goal of the present work was to find 
a compression scheme for a given deep neural model. Inspired by techniques used in coresets, we experimented with the farthest-first (FF) traversal, also known as 
the Gonzalez algorithm \citep{gonzalez1985clustering}, which can be used 
to obtain an efficient 2-approximation algorithm for the $k$-center clustering problem \citep{hochbaum1996approximation}.
Given a set of points in a metric space, its 
FF traversal is constructed by taking the 
first point $x$ arbitrarily, then take the farthest point from $x$ as the next point, and in subsequent steps
always greedily choose the point farthest away from any of the points already chosen.

When dealing with complex input signals such as images or sound, it makes little sense
to consider the input space itself. A natural observation is that in a trained 
deep model, there are representation layers that create 
manifolds on which \emph{semantically} similar objects tend to be closer to each 
other. Thus, the geometry over spaces 
induced by these layers can be useful for creating
coresets.

The basic FF idea thus gives rise to the following compression algorithm, which we call \emph{farthest-first compression} (FF-Comp). 
Consider a multi-class classification problem with $k$ classes.  Given a training set, $S_m$, we train a deep neural model
$f(x)=\tau(\phi(x))$, where $\phi(\cdot)$ represents the the entire network
excluding the last layer (see Section~\ref{sec:setting}).
We construct $k$ coresets, one for each class, 
using FF traversals over the spaces
$S'_{m,i} = \{ \phi(x) \ : \ (x,y) \in S_m, \ y = i \}$.
Formally,  suppose we are creating the coresets $C_i$, $i = 1,\ldots,k$.
Denoting $d(u,v) = \| u - v \|_2$, for a non-empty $C_i$, the next labeled point, $(x',y') \in S_{m,i}$, to be added to $C_i$, is 
\begin{eqnarray}
\label{eq:ff}
(x',y')&=&\argmax_{(x',y')\in S_m} \min_{x:(x,y)\in S_m} d(\phi(x),\phi(x'))\\
s.t&& y'=i\nonumber.
\end{eqnarray}
The resulting compression algorithm is given in  
Algorithm 1. This algorithm essentially selects up to 
$c$ points from 
$S_m$ in a ``stratified'' manner.

\begin{algorithm}[htb]
	\label{alg:compress}
	\caption{\emph{Farthest-First Compression} (FF-Comp)}
	\begin{algorithmic}[1]
		\STATE FF-Comp($S_m$,$\phi(\cdot)$,$c$,$k$)
		\FOR {$i = 1$ \TO $k$}
		\STATE draw a random seed $(x,y) \in S_m:y=i$
		\FOR {$j = 1$ \TO $\lfloor c/k \rfloor$}
		\STATE find $(x',y')$ according to Equation~((\ref{eq:ff})
		\STATE $S'_c \leftarrow (x',y')$
		\ENDFOR
		\ENDFOR
		\STATE  Output- $S_c$
	\end{algorithmic}
\end{algorithm}

To evaluate the performance of the resulting compression, we retrain
the same architecture using $S_c = \cup_i C_i$, and assess its test error  over
an independent labeled set.
Applying FF-Comp over the CIFAR-10 dataset, we obtained 50\% compression
with a degrading test error of only 1.5\% (93.23\% accuracy before, and  
91.73\% after). While this might seem to be a very impressive compression,
comparing it to the 3.1\% accuracy reduction obtained
by a random 50\% compression is somewhat disappointing.
It turns out, however, that the required random sub-sampling rate that will lead to 
91.73\% accuracy over the test set (as in the more clever FF compression) is 64\%, which amounts to an
additional 7,000 labeled training points to match the compression 
performance. Viewed from an active learning perspective,
this is a large saving that can potentially be exploited.  This result motivates the construction of a new active learning algorithm based on FF traversals over the representation level of a \emph{trained} model within the proposed ``long tail'' setting of active learning.


\section{Deep Active Learning with Coresets}
\label{sec:active_deep}

The compression result of Section~\ref{sec:compression} motivates 
an active learning algorithm whose querying function operates by
computing coresets. 
We consider the long-tail active setting of Section~\ref{sec:setting} whereby
we already have a trained deep model $f(x)=\tau(\phi(x))$ that was trained over
$S_m$. We also assume we have access to $S_m$ itself,
as well as to a pool
$U$ of unlabeled points.
At each stage $t$ in the active session we have 
a labeled training set $L_t$ and an unlabeled pool
$U_t$ (initially, $L_0 = S_m$ and $U_{0} = U$) and we would like to select additional 
$b$ points from $U_{t-1}$, request their label, and then re-train 
$f$ over $L_t$ (which is the union of $L_{t-1}$ with the newly
acquiried $b$ labeled points).

We would like to apply the same coreset principle as in the FF-Comp compression
scheme. However, note that in the active game, the pool $U_t$ is unlabeled and
we cannot stratify the queried points. 
Algorithm 2 provides the pseudo-code for our active learning algorithm,
for which we intentionally formulated only the basic principle
without applying various potential improvements such as
pseudo labeling (which can be used to apply stratification or to enrich the labeled training set at each iteration).
The algorithm receives as input the initial classifier $f_0$,
its training set $S_m$, the unlabeled pool $U$, and 
a batch size $b$, defined to be the 
number of extra pool points to be queried at each round.

\begin{algorithm}[htb]
	\label{alg:activeff}
	\caption{\emph{Farthest First Active Learning} (FF-Active)}\label{CC}
	\begin{algorithmic}[1]
    \STATE FF($S_m$,$U$,$f_0$,$b$)
	    \STATE $L_0=S_m$
        \STATE t=0
		\WHILE {budget not exceeded / desired accuracy have not met}
                \STATE $t=t+1$
                \STATE $S_b=\emptyset$
        \FOR {$i = 1$ \TO $b$}
		\STATE $S_b = S_b\cup \argmax_{(x',y') \in U}(\min_{(x,y)\in L_{t-1}\cup S_b}(d(\phi(x'),\phi(x))))$
		\ENDFOR
        \STATE $L_{t}=L_{t-1} \cup S_b$
       
        \STATE  train  $f_t$ using $L_t$
		\ENDWHILE
		\STATE  Output- $f_t$
	\end{algorithmic}
\end{algorithm}

 .

\section{Experiments}
\label{sec:experiments}
In this section we report on the results of several experiments. In each experiment we compare the performance 
of our FF-Active algorithm to that of the traditional softmax response (SR) method (uncertainty sampling) and to Random (passive learning). We experimented with three standard 
datasets: MNIST, CIFAR-10, and CIFAR-100.
These experiments indicate that FF-Active has significant advantage over Random, and that it is 
better than SR.
We also present an experiment over a synthetic expansion of 
CIFAR-100, which highlights a more challenging scenario for 
SR.

\subsection{MNIST}
\begin{figure}[htb]
	\centering
		\includegraphics[width=0.6\linewidth,height=0.4\linewidth]{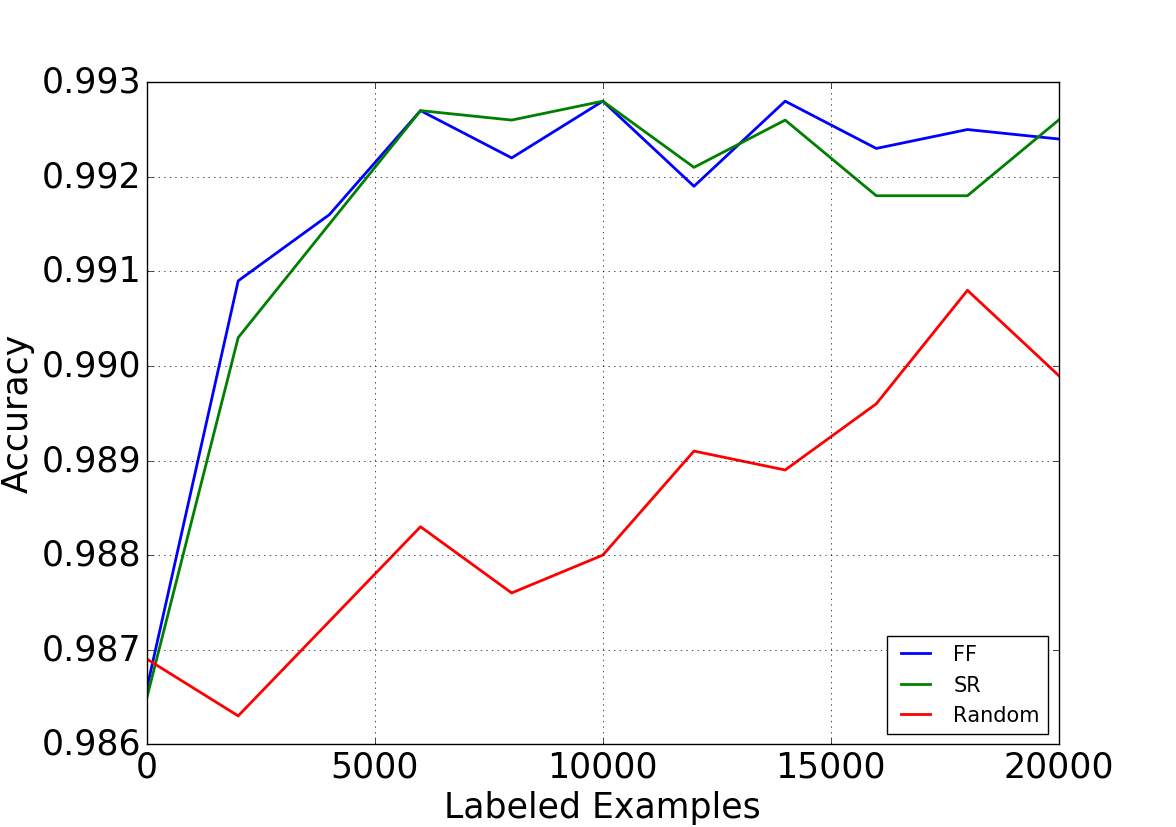}\label{fig:Mnist}
	\caption{Test accuracy as function of number of points labeled for MNIST dataset.}
  \end{figure}

We begin by testing FF-Active over the MNIST dataset
for which we trained a network similar to LeNet \citep{lecun1998gradient}, 
whose architecture contains two convolutional layers, one fully-connected hidden layer and a softmax layer. We used Adam as the optimization algorithm. 
MNIST consists of 10 classes of images of hand-written digits, 
and contains 60,000 labeled examples.
We performed the initial training over a labeled set, $S_m$, containing 
10,000 images sampled uniformly at random from the
entire set. The remaining 50,000 images were taken to be the unlabeled pool $U$.
The active session consists of rounds
where in each one, each active learning algorithm selected an additional 2000 points
from the pool according to its querying function.
The resulting learning curves of FF-Active, SR and Random are presented
in Figure~\ref{fig:Mnist}. It is evident that 
FF-Active quickly extracted much of the relevant information in $U$ after 8000
additional labeled points. Random, on the other hand, did not achieve this performance
level even after 20,000 additional labeled queries.
In this dataset the performance of SR is indistinguishable from that of FF-Active.

\subsection{CIFAR-10}
\label{sec:cifar-10}
\begin{figure}[htb]
	\centering
		\includegraphics[width=0.6\linewidth,,height=0.4\linewidth]{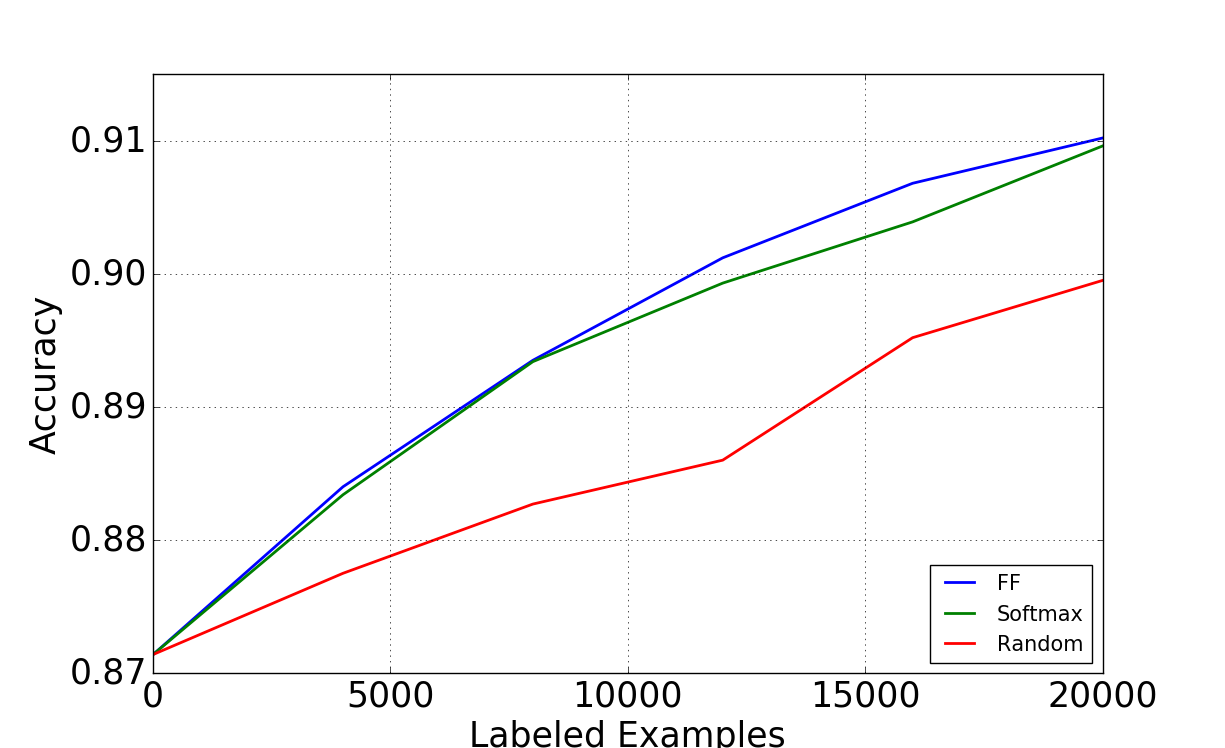}\label{fig:cifar10}
	\caption{Test accuracy as function of number of points labeled for CIFAR-10 dataset.}
  \end{figure}

For the CIFAR-10 dataset \citep{krizhevsky2009learning} we employed the
VGG-16 architecture \citep{simonyan2014very}.
We trained the model for 250 epochs using \emph{momentum stochastic gradient descent} (SGD) with a batch size of 128 and a learning rate of 0.1 and multiplicative learning rate drop of 0.5 every 20 epochs. These hyper-parameters were selected using the initial training set $S_m$, consisting of 25,000 images that were chosen uniformly at random
from the entire set (50,000 images).
The remaining 25,000 images where used for the pool $U$ that was doubled by
horizontal flips\footnote{These horizontal flips were not used to augment 
$U$ in the case of MNIST because they do not represent valid digits.}
In each active round, the algorithm selected additional 4000
images from $U$ to be labeled. 
Figure~\ref{fig:cifar10} presents the learning curves of 
FF-Active, SR and Random. We observe that FF-Active is substantially more 
label efficient than Random and its advantage even increases through the active learning session.
The performance of SR is nearly identical to that of 
FF-Active through the first 8000 additional examples, and then deteriorates.

\subsection{CIFAR-100}
\label{sec:cifar100}
\begin{figure}[htb]
	\centering
		\includegraphics[width=0.6\linewidth,,height=0.4\linewidth]{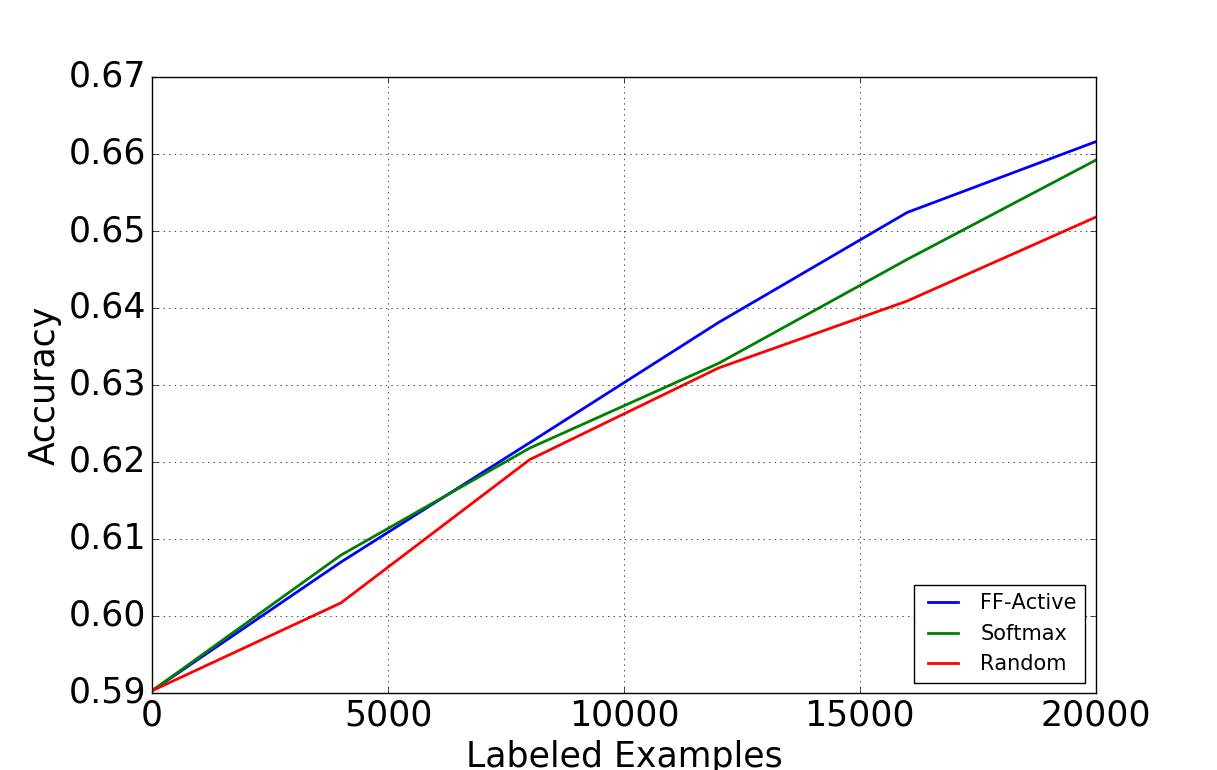}\label{fig:cifar100}
	\caption{Test accuracy as function of number of points labeled for CIFAR-100 dataset.}
  \end{figure}

For CIFAR-100 we used an identical VGG-16 architecture to that used for 
CIFAR-10 (with the exception of the last output layer that now consisted
of 100 units). 
We also used the same optimization algorithm and learning rate schedule. 
The construction of $S_m$ and $U$ was also identical to the CIFAR-10
experiment. 
The learning curves of FF-Active, SR, and Random are depicted in Figure~\ref{fig:cifar100}.
Here again we see a consistent advantage of FF-Active over Random and near identical performance of FF-active and SR during the
initial stage and then degradation of SR.
While the relative advantage of FF-Active
over Random appears to be smaller in CIFAR-100 compared to CIFAR-10,
 the overall accuracy slope (accuracy improvement vs. 
 labeled examples) is larger in CIFAR-100 than 
 in CIFAR-10.
 Also, we observe that the gap between FF-Active and Random 
 is increasing (in both CIFAR-100 and CIFAR-10).
It would, therefore, be very interesting to 
examine what the result would be using a very large pool
(a really long tail); hence, our next experiment.

\subsection{The Longer Tail -- Synthetically Inflated Pool}
Our active learning setting is designed to model
potential label complexity saving over a prolonged 
active session where more and more examples are available for
labeling. While we expect this scenario to occur 
in future applications of machine learning models,
currently available datasets aren't large enough to model this scenario. 

For the experiments in this section, we synthetically created a larger dataset, based on a CIFAR-100, attempting to 
model a larger pool. Specifically, we inflated
the CIFAR-100 pool by a factor of three, using 
bootstrap sampling (sampling with replacement).
We experimented with the resulting inflated CIFAR-100
using the same experimental setting described in Section~\ref{sec:cifar100}. The unlabeled pool $U$ in this experiment contains 150,000 images.
Here we observe a very significant domination 
of FF-Active over Random, and domination of 
FF-Active over SR through most of the session.

\begin{figure}[htb]
	\centering
    \includegraphics[width=0.6\linewidth,height=0.4\linewidth]{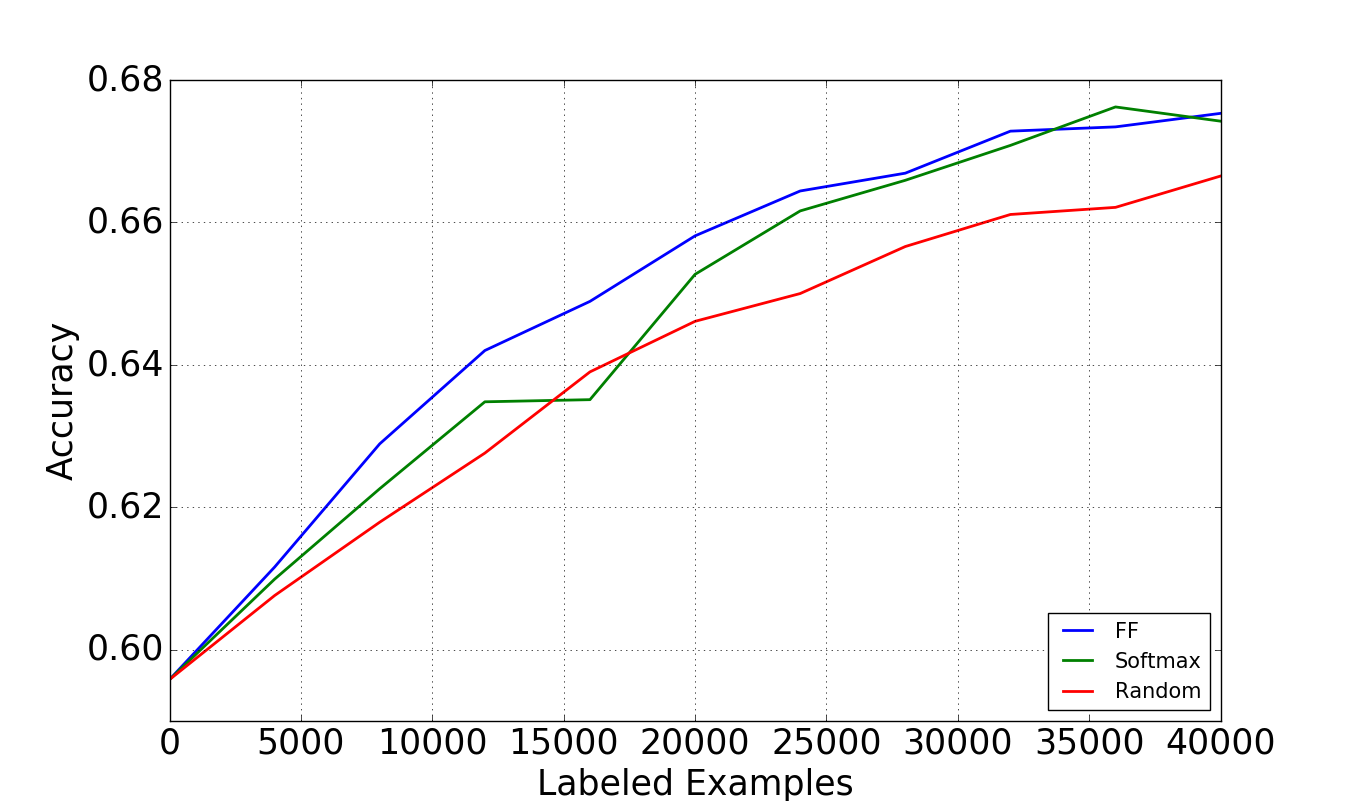}\label{fig:extreme}
	\caption{Test accuracy as a function of the number of points labeled. }
\end{figure}

\section{Inspecting of FF and SR over a Small Synthetic Example}
\begin{figure}[htb]
\label{fig:simulation}
	\begin{center}
		\subfigure[Problem setting]{\includegraphics[width=0.3\linewidth]{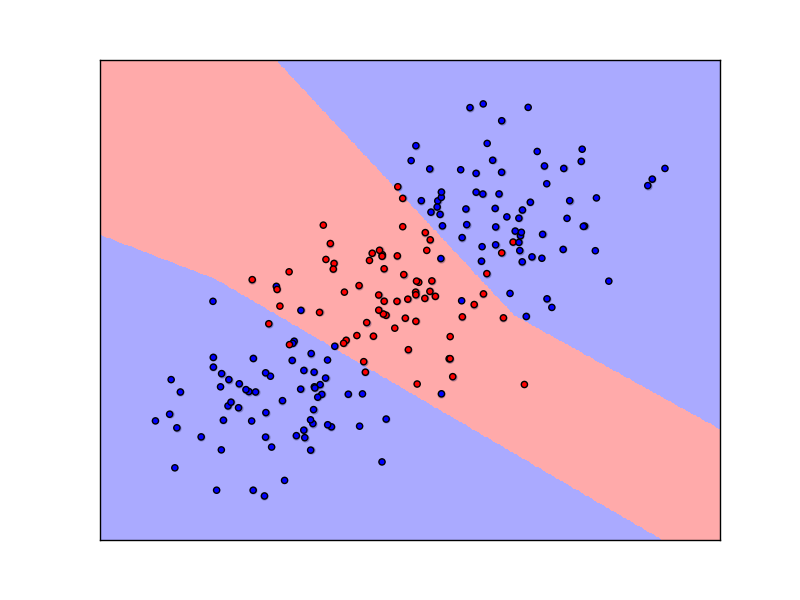}\label{fig:problem}}
        \subfigure[Initialization]{\includegraphics[width=0.3\linewidth]{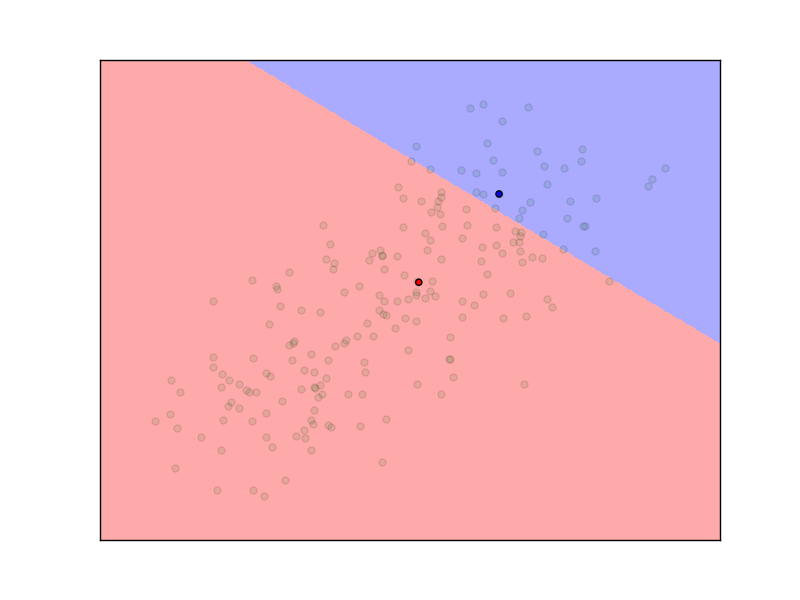}\label{fig:stage1}}

        \subfigure[SR - 6 points]{\includegraphics[width=0.3\linewidth]{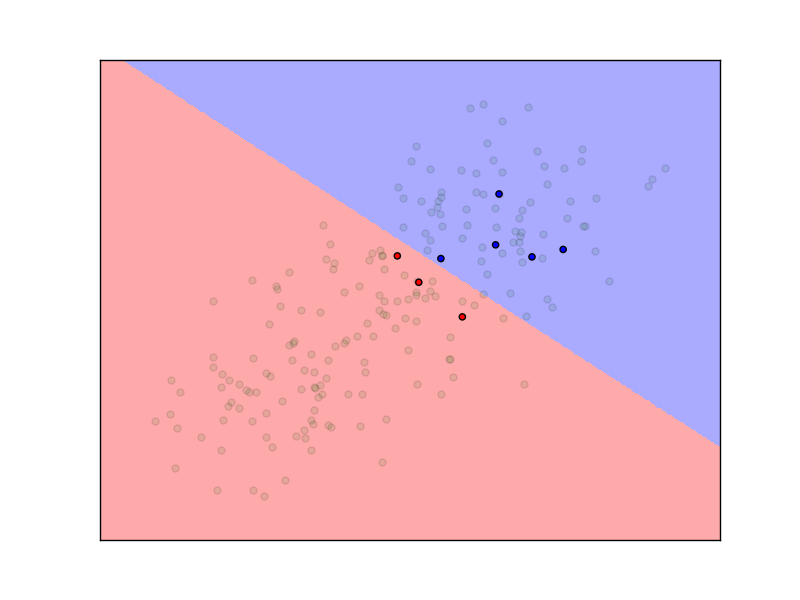}\label{fig:1}}
        \subfigure[SR - 15 points]{\includegraphics[width=0.3\linewidth]{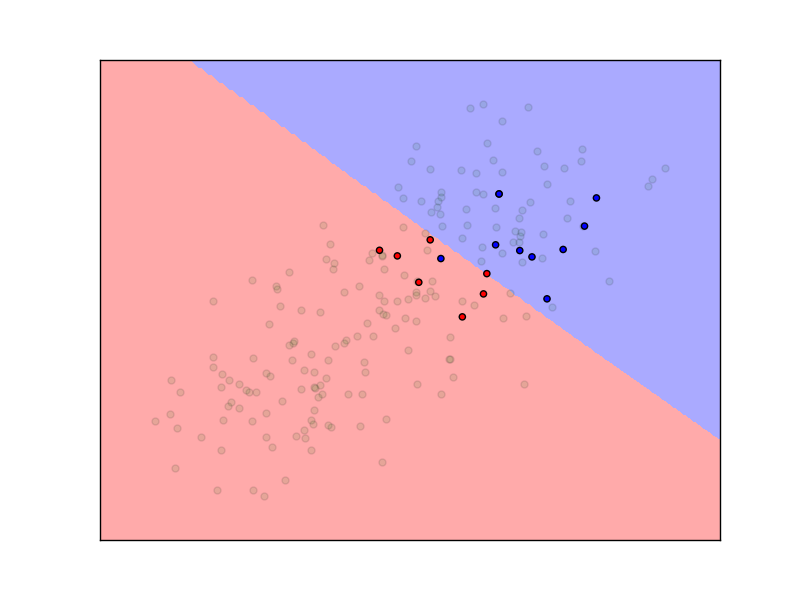}\label{fig:3}}
        \subfigure[SR - 30 points]{\includegraphics[width=0.3\linewidth]{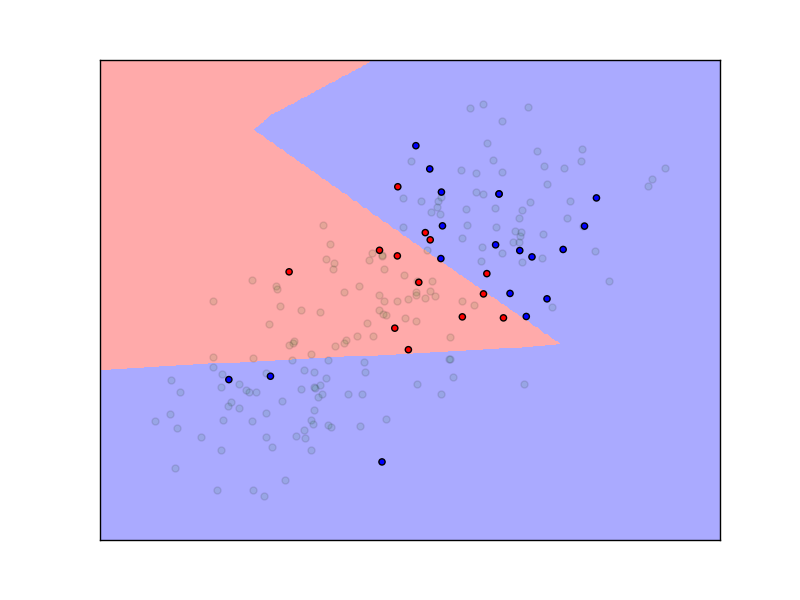}\label{fig:5}}

               \subfigure[FF - 6 points]{\includegraphics[width=0.3\linewidth]{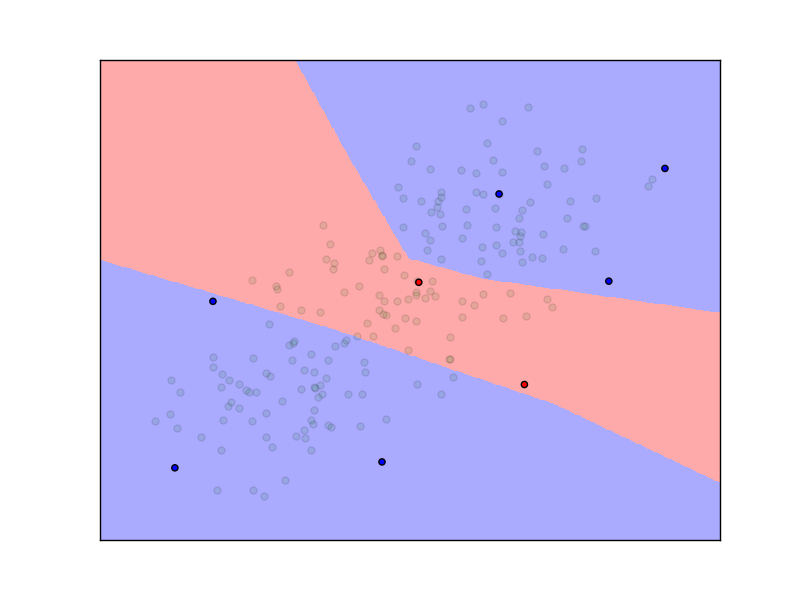}\label{fig:1}}
        \subfigure[FF - 15 points]{\includegraphics[width=0.3\linewidth]{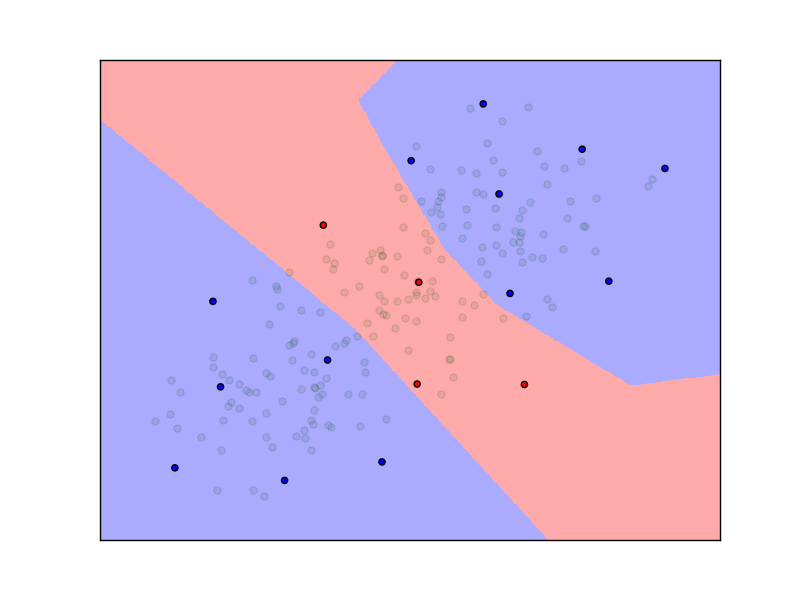}\label{fig:3}}
        \subfigure[FF - 30 points]{\includegraphics[width=0.3\linewidth]{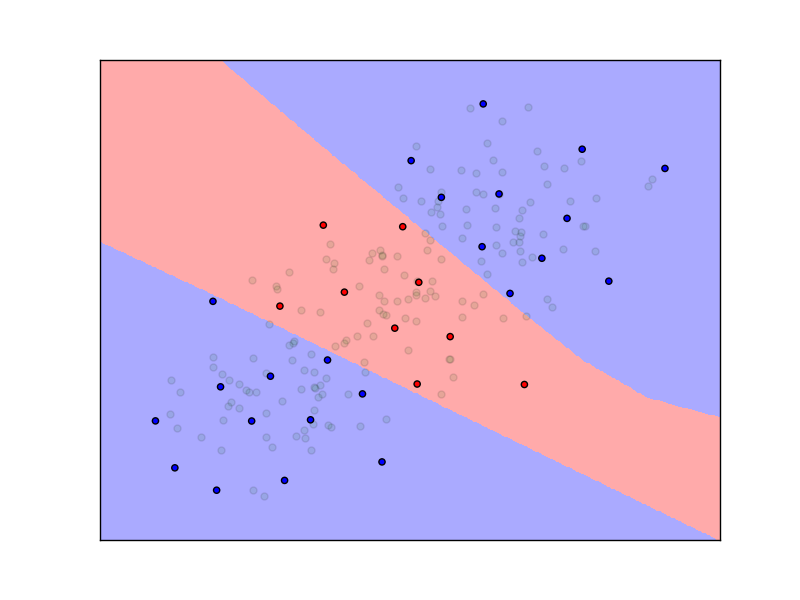}\label{fig:5}}
      \end{center}
	\caption{Simulating FF-Active and SR over a 2D example.}
	\label{fig:fff}
\end{figure}

In this section we observe more closely the 
behavior of the querying functions of FF-Active and SR,
and demonstrate that their strategies are quite different.
The FF-Active strategy conducts systematic exploration in the sense that its next query is always on the point least represented by the current coreset (using the metric we choose to employ in the underlying representation space). On the other hand, SR is focused on exploiting (refining) the region around the decision boundary.
We demonstrate this difference by simulating the methods over a synthetic dataset representing the binary learning problem 
depicted in Figure~\ref{fig:problem}. In this figure there are 200 points randomly sampled from three Gaussians in $\reals^2$.  The middle Gaussian represents the negative class (red) and the two other Gaussians represents the positive class (blue). Assume that we randomly received two labeled points, one in each class, to initialize an active learning session for both 
FF-Active and SR. All other points at this stage consist the unlabeled pool (depicted  in faded green). A neural network with one hidden layer is in use, in Figure~\ref{fig:stage1} where the initial classifier is depicted together with the two initial labeled points.
We now independently apply the two algorithms (FF-Active and SR) starting from this initial state. At each round we compute the next query, receive its label, and revise the decision boundary by re-training the model with the 
revised training set.
Figure~\ref{fig:simulation} depicts the results of the two simulations.  It is evident that FF-Active effectively captures the geometry of the problem using much fewer queries than SR.

This simulation nicely illustrates the conceptual difference
between the two strategies. Whereas FF-Active was able 
to almost perfectly identify the best model with 6 queries, SR is still far away with 30 queries.
The caveat here is that this simulation is focused on an early stage of the active learning process, and we are interested in the later stages (the ``long tail'').
We speculate that in complex problems such as CIFAR-10 and CIFAR-100 (noisy, high dimensional, and multi-class),  extensive exploration is required throughout the game over the representation space 
created by the DNN.
We leave this interesting question
for future research.



\section{Concluding Remarks}
Previous studies indicate that without   
prior knowledge or
hindsight, active learning cannot be
be effectively
performed.
In this paper we focus on a setting where an initial 
reasonable model has already been trained and only then
 we start to learn actively.
Our results indicate that considerable labeling resources can be saved using an active algorithm whose goal is to improve this initial model.

Motivated by compression ideas, our main contribution is a novel pool-based active learning 
algorithm for deep nets achieving clear and significant advantage over passive learning. The traditional softmax-response (SR) technique
that has been previously considered for deep active
learning is also useful in this setting, but
is inferior to our method.
Overall, we believe that the proposed method provides a viable and 
practical tool that will work `out-of-the-box' on
image data and possibly on other types of data.

The main idea behind our new querying function is the use of model-based coresets for compressing the input data 
based on its internal representation over the space defined 
by neuronal activations over a representation layer. 
The specific coreset engine in our case is the greedy
farthest-first traversal. Many other interesting and perhaps 
more powerful coreset engines can be considered,
such as
large margin coresets \citep{har2007maximum},
coresets used for DP-means clustering \citep{bachem2015coresets}, and
coresets developed for dimensionality reduction \citep{feldman2016dimensionality}, to name a few.
Of course, by adapting the coreset view, it would be 
very interesting to prove approximation guarantees 
for neural networks using existing or new techniques.

Other improvements of the proposed method would be 
very interesting to consider. For example,
the use pseudo-labeling can allow for stratified coresets,
and also enable inferring true labels using high confidence principles
\citep{Nips2017}. We also believe that 
using the Monte-Carlo dropout technique of
\citet{gal2017deep}, we can effectively generate an ensemble of 
representations and reduce the variance in our traversal predictions, possibly decreasing the label complexity.

Finally, we would like to emphasize the open question 
whose study motivated this work. Would it be possible to 
compress datasets such as CIFAR-100 or Imagenet to a (logarithmic)
fraction of their size while maintaining high 
classification performance?
Any significant advancement in this direction is likely to 
substantially advance deep active learning and moreover, 
can potentially establish a theoretical breakthrough in the theory of deep learning 
using compression scheme techniques of statistical learning theory \citep{shai_moran}.


\section*{Acknowledgments}
This research was supported by The Israel Science Foundation (grant No. 1890/14)

\bibliography{iclr2018_conference}
\bibliographystyle{iclr2018_conference}

\end{document}